\documentclass[11pt]{article}
\usepackage[utf8]{inputenc}
\usepackage[T1]{fontenc}
\usepackage{geometry}
\usepackage{booktabs}
\usepackage{hyperref}
\usepackage{amsmath}
\usepackage{amssymb}
\usepackage{microtype}
\usepackage{hyphenat}
\makeatletter
\renewcommand\paragraph{\@startsection{paragraph}{4}{\z@}%
  {1.0ex plus .2ex minus .2ex}%
  {-0.8em}%
  {\normalfont\small\bfseries}}
\makeatother

\title{Verification and Identification in ECG biometric on large-scale.} 
\author{Scagnetto A.\\ASUGI - Azienda Sanitaria Universitaria Giuliano-Isontina}
\date{}

\begin{document}
\maketitle

\begin{abstract}
This work studies electrocardiogram (ECG) biometrics at large scale, directly addressing a critical gap in the literature: the scarcity of large-scale evaluations with operational metrics and protocols that enable meaningful standardization and comparison across studies.

We show that identity information is already present in tabular representations (fiducial features): even a simple MLP-based embedding network yields non-trivial performance, establishing a strong baseline before waveform modeling.

We then adopt embedding-based deep learning models (ArcFace), first on features and then on ECG waveforms, showing a clear performance jump when moving from tabular inputs to waveforms, and a further gain with larger training sets and consistent normalization across train/val/test.

On a large-scale test set, verification achieves high TAR at strict FAR thresholds (TAR=0.908 @ FAR=1e-3; TAR=0.820 @ FAR=1e-4) with EER=2.53\% (all-vs-all); closed-set identification yields Rank@1=0.812 and Rank@10=0.910. In open-set, a two-stage pipeline (top-$K$ shortlist on embeddings + re-ranking) reaches DIR@FAR up to 0.976 at FAR=1e-3 and 1e-4.

Overall, the results show that ECG carries a measurable individual signature and that large-scale testing is essential to obtain realistic, comparable metrics. The study provides an operationally grounded benchmark that helps standardize evaluation across protocols.
\end{abstract}

\section{Introduction}

The electrocardiogram (ECG) is a physiological signal rich in morphological and temporal information, influenced by clinical state, acquisition conditions, and inter-session variability. Despite this variability, extensive experimental evidence and literature reviews indicate an individual component that is stable enough to support biometric verification (1:1) and identification (1:N). In particular, classic and recent works agree on a statistically and operationally relevant separation between intra-subject (genuine) and inter-subject (impostor) comparisons, with identity information persisting over months or years, albeit with performance degradation compared to single-session scenarios \cite{odinaka2012,donida2019,fratini2015,fatemian2009,dasilva2014}.

The literature highlights two recurring critical issues that limit comparability across studies and the operational interpretation of results: (i) the need for clear and comparable experimental protocols (dataset choice, split criteria, session handling, acquisition conditions); and (ii) the adoption of metrics consistent with operational biometrics rather than purely ``laboratory'' settings, capable of describing system behavior at realistic security thresholds \cite{odinaka2012,donida2019,fratini2015}. A third, practical gap is the scarcity of large-scale evaluations, which prevents meaningful standardization of metrics and protocols. On this basis, this study analyzes ECG biometrics at large scale, using a test set of about 20,000 identities and reporting metrics and protocols that make performance interpretable in real-world applications.

In recent years, the adoption of deep learning models and embedding representations has accelerated ECG biometric systems, but heterogeneous experimental settings (dataset size, preprocessing, multi-session, and metric choice) still make direct comparison difficult. In this context, the contribution of this work is primarily methodological and experimental: rather than proposing a new loss or architecture, we provide large-scale evidence with operational metrics to promote standardization and to build and evaluate a reproducible, scalable, and deployment-oriented pipeline.

Specifically, this study: (i) evaluates an embedding-based system at large scale (tens of thousands of identities) using operational biometric metrics (Rank@K, TAR@FAR, EER), including operating points with very low FAR; (ii) extends evaluation to open-set via DIR@FAR in a two-stage pipeline (top-$K$ shortlist on embeddings + re-ranking/decision), closer to watchlist scenarios; (iii) estimates genuine/impostor separability robustly via all-vs-all pairing, reducing optimistic bias as population size grows; (iv) explicitly links deep performance to preliminary feature-based evidence (INTRA > INTER) within a defined temporal window; (v) quantifies the impact of under-reported choices in the literature (waveform vs tabular representation, per-sample vs global normalization, training set composition); (vi) interprets identification results in terms of a two-stage pipeline (candidate generation followed by re-ranking/verification), consistent with realistic deployment requirements.

The study proceeds in two logical phases:
\begin{enumerate}
\item \textbf{Statistical motivation:} verify, on ECG tabular features, that similarity between two exams of the same patient is systematically higher than similarity between exams from different patients, quantifying the ``subject effect'' before deep learning training \cite{merone2017,pinto2018,melzi2023,meltzer2025,goldberger2000}.

\item \textbf{Biometric evaluation with deep learning:} learn embeddings and measure performance on a patient-disjoint test set with metrics suitable for operational biometrics (TAR@FAR, EER, Rank@K), comparing two representations: tabular fiducial features (axes, durations, onset/offset: 13 variables) and ECG waveforms.
\end{enumerate}

\section{Materials and Methods}

\subsection{Datasets}

The data come from a clinical database of the Azienda Ospedaliera Universitaria Giuliano-Isontina (Friuli-Venezia-Giulia, Italy). The repository contains about 1 million ECGs both as fiducial \emph{features} (computed by Mortara algorithms) and as multichannel \emph{waveform} signals over 12 synchronized leads (duration 10 s, sampling 1 kHz, 16-bit quantization).

For the analyses in this work we used only exams with valid \texttt{PatientID} (non-missing and non-alphanumeric), totaling 687,147 ECGs from 191,355 patients. From this base we apply three refinement steps to reduce technical variability and outliers while preserving a large cohort:
\begin{enumerate}
\item \textbf{Multi-exam selection and temporal constraint:} patients with more than 2 exams and distance $>30$ days between consecutive exams (395,095 exams, 92,523 patients).
\item \textbf{Outlier cleaning and cap per patient:} outlier removal via extended physiological ranges (Table~\ref{tab:range}) and a maximum of 10 exams per patient (168,999 exams, 54,218 patients).
\item \textbf{Device homogenization:} select only ECGs acquired with ELI250 device (164,440 exams, 53,079 patients).
\end{enumerate}

The resulting cohort is the common base for both the statistical and operational pipelines. For each pipeline step, patient frequency by number of exams is also reported (Table~\ref{tab:steps}).

\subsubsection{Dataset for correlation statistics}

For the INTRA vs INTER analysis, a dedicated dataset is built by selecting, for each patient, two exams spaced 6--18 months apart. A months-scale distance makes the comparison more challenging (greater intra-subject variability) and allows a more robust test of a persistent ``signature''.

On this dataset we build: (i) \textbf{INTRA} pairs (one pair per patient, between the two selected exams) and (ii) \textbf{INTER} pairs (same number of pairs, obtained by pairing exams from different patients with a fixed seed). For each pair, Pearson and Spearman correlations are computed on a set of ECG \emph{features}.

Note: correlation statistics use 8 features (not 13) to avoid redundancy, since 5 of the 13 are linear combinations of the other 8.

\subsubsection{Dataset for deep learning}

For embedding-based models we use a different temporal constraint: patients with at least 2 exams, cap 2--10 exams per patient, and minimum distance of 30 days between consecutive exams (period 2006-02-20 \textrightarrow{} 2020-12-31). On this cohort we define the tabular representation for Siamese and ArcFace by selecting columns \texttt{PatientID}, \texttt{ExamID}, \texttt{AcquisitionDate}, \texttt{Gender}, \texttt{PatientAge}, and 13 fiducial features (VentricularRate, PRInterval, QRSDuration, QTInterval, QTCorrected, PAxis, RAxis, TAxis, QOnset, QOffset, POnset, POffset, TOffset).

Three datasets were generated with different Train/Validation/Test splits; statistics are reported in Tables~\ref{tab:train} (train), \ref{tab:val} (validation), and \ref{tab:test} (test). In all cases, splitting is performed \emph{by patient} (no subject shared across train/val/test) to prevent leakage. The three datasets also differ by the minimum number of exams per patient in train (3/4/5), which helps stabilize training with contrastive/triplet losses that require informative pair/triplet mining (genuine/impostor) \cite{schroff2015,hermans2017}.

\subsection{INTRA vs INTER Correlation Study}

This analysis aims to verify, in a controlled manner and prior to deep model training, whether similarity between two exams of the same subject is systematically higher than similarity between exams from different subjects.

\paragraph{Tabular representation.} For each ECG exam we build a vector of fiducial features ($d=8$ in the first analysis). Similarity between two exams $x_i$ and $x_j$ is measured by Pearson and Spearman correlation between their feature vectors.

\paragraph{Pair construction.} Pairs are defined on a dedicated dataset where, for each patient, two exams 6--18 months apart are selected. We consider:
\begin{itemize}
\item \textbf{INTRA:} one pair per patient between the two selected exams (genuine comparisons);
\item \textbf{INTER:} the same number of pairs, obtained by pairing exams from different patients (impostor comparisons) with a fixed seed.
\end{itemize}
This procedure balances INTRA and INTER comparisons and makes impostor sampling reproducible.

\paragraph{Statistical analysis.} We compare INTRA and INTER correlation distributions (Pearson and Spearman) using non-parametric tests and robustness analyses (bootstrap/effect size), to quantify separation between the two comparison classes.

Results (INTRA vs INTER separation and comparison with literature) are reported in Section~\ref{sec:risultati_stat}.

\subsection{Verification and Identification Study}

Verification (1:1) and identification (1:N) tasks are the standard evaluation in biometrics. In closed-set, every probe belongs to an identity present in the gallery (no unknown classes), whereas in open-set some probes may belong to identities not present in the gallery and the system must reject unknown identities before producing a ranking. In verification, the comparison is between probe and the declared identity template (accept/reject), with typical metrics TAR@FAR, EER, and ROC/DET curves; in identification, the probe is compared to the entire gallery, with metrics Rank@K (CMC) and top-1 accuracy. In open-set settings, specific metrics (e.g., DIR@FAR) and rejection thresholds are used. In this work, all reported performance is computed in closed-set unless explicitly stated.

\paragraph{Open-set (DIR@FAR) and second stage.} Introducing a second stage of re-ranking/decision on a top-$K$ shortlist necessarily operates in open-set: some probes may not belong to the gallery and must be rejected at threshold. We use a two-stage pipeline: (i) top-$K$ shortlist on ArcFace embeddings with cosine similarity, (ii) re-ranking/decision on the top-$K$ candidates with three score aggregation strategies: \emph{best-of-$K$} (max-score fusion) \cite{kittler1998}, \emph{top-$K$ mean} (mean-score fusion) \cite{kittler1998}, and \emph{s-norm} \cite{auckenthaler2000}. Open-set evaluation uses DIR@FAR, with thresholds calibrated on impostors and acceptance of known probes only if the score exceeds the threshold \cite{poh2012}.

\subsubsection{Verification: TAR@FAR and EER}

In operational biometrics, an operating point is chosen by fixing a false accept risk (FAR, False Acceptance Rate) and measuring the achievable sensitivity (TAR, True Acceptance Rate = 1 - FRR, False Rejection Rate) under that constraint. TAR@FAR is thus often more informative than accuracy and, in many scenarios, more useful than AUC (Area Under the Curve) \cite{chee2022}. In a verification setting, TAR@FAR presupposes a declared identity or an enrollment template in the gallery: probes without a corresponding template are not part of the metric calculation because genuine/impostor pairing is undefined.
EER (Equal Error Rate) is the ROC (Receiver Operating Characteristic)/DET (Detection Error Tradeoff) point where FAR = FRR: a global summary of separation between genuine and impostor scores, useful for quick comparisons \emph{under the same protocol}.

In this study, EER is estimated robustly: all-vs-all over all probe-gallery pairs in the test set (upper triangle), with blockwise similarity counting and identification of the FAR=FRR point. This avoids optimistic bias from impostor sampling as population grows.

\subsubsection{Identification: Rank@K (CMC) and ``accuracy''}

Rank@K measures the probability that the correct identity appears among the top K candidates. Rank@1 coincides with top-1 identification accuracy only in a closed-set setting and with a defined scoring/ranking protocol; many publications use ``accuracy'' ambiguously (classification accuracy on seen subjects, accuracy on beats/segments, or top-k), making comparisons non-trivial \cite{merone2017,pinto2018,melzi2023}.

\subsection{Networks}
We developed three models, or rather three approaches to the ``VerIde'' (VERification IDEntification) problem: a Siamese MLP for contrastive and triplet losses on tabular data, and two ArcFace approaches, one for tabular variables with an MLP backbone and one for waveform signals with a convolutional backbone.

\subsubsection{Siamese}
The Siamese MLP backbone takes as input 13 fiducial features. The structure is a sequence of fully-connected layers with dimensions 32-64-256, each with BatchNorm1d and ReLU (Rectified Linear Unit), followed by dropout 0.10. The output is a 128-dimensional embedding obtained via a final linear layer; the embedding is then L2-normalized (Euclidean norm) to use distances/cosine as similarity measures.

Siamese networks are natural for ``same/different'' problems: two branches with shared weights map two inputs to embeddings $z_1=f(x_1)$ and $z_2=f(x_2)$ and directly optimize a similarity metric \cite{bromley1993,chopra2005,hadsell2006}. Unlike multi-class classification, the loss acts on the metric space and enables the same embedding for verification (thresholding a score) and identification (ranking within a gallery).

\paragraph{Contrastive loss (pairs).} We consider a pair $(x_1, x_2)$ and a label $y\in\{0,1\}$, where $y=1$ indicates same identity (genuine) and $y=0$ different identities (impostor). Given a distance in embedding space, e.g.,
\[
  D = \lVert z_1 - z_2 \rVert_2,
\]
the contrastive loss penalizes genuine pairs if they are too far and impostor pairs if they are too close. A standard form is \cite{hadsell2006}:
\[
  \mathcal{L}_{\text{contr}} = y\,D^2 + (1-y)\,\max(0, m-D)^2,
\]
where $m>0$ is a margin. Intuitively: (i) genuine pairs are compressed (reducing $D$), (ii) impostors are pushed at least beyond $m$ (zero loss for $D\ge m$). The margin controls the minimum required separation between different identities.

\paragraph{Triplet loss (triplets).} We consider a triplet $(x_a, x_p, x_n)$: anchor $x_a$, positive $x_p$ (same identity as anchor), and negative $x_n$ (different identity). With $z_a=f(x_a)$, $z_p=f(x_p)$, $z_n=f(x_n)$ and distances $D_{ap}=\lVert z_a-z_p\rVert$ and $D_{an}=\lVert z_a-z_n\rVert$, the triplet loss enforces a relative constraint \cite{schroff2015}:
\[
  \mathcal{L}_{\text{triplet}} = \max\bigl(0,\; D_{ap} - D_{an} + \alpha \bigr),
\]
with margin $\alpha>0$. Here the goal is not just to ``pull'' or ``push'' in absolute terms, but to ensure the negative is at least $\alpha$ farther than the positive: $D_{an} \ge D_{ap}+\alpha$. In practice, optimization quality depends on triplet \emph{sampling} (random, semi-hard, hard), because very easy triplets give zero loss and do not update weights \cite{hermans2017}.

\subsubsection{ArcFace}

ArcFace introduces the \emph{Additive Angular Margin Loss}: a formulation where both classifier weights and embeddings are normalized so that the decision depends only on the angle between vectors, not their norms \cite{deng2019}. The architecture is split into a backbone (embedding extractor) and a head (classifier with angular margin): during training, multi-class classification is optimized; after training, the head is removed and only the backbone is used to obtain embeddings comparable via cosine similarity (verification) or ranking over a gallery (identification).

\paragraph{Backbone and embedding.} The backbone produces an embedding $x\in\mathbb{R}^d$ that is normalized:
\[
  \hat{x} = \frac{x}{\lVert x \rVert}.
\]
In parallel, in the classification layer each class $j$ has a weight vector $W_j\in\mathbb{R}^d$ which is also normalized:
\[
  \hat{W}_j = \frac{W_j}{\lVert W_j \rVert}.
\]
The logit (pre-softmax) becomes proportional to cosine similarity between embedding and class weight:
\[
  \hat{W}_j^T \hat{x} = \cos(\theta_j),
\]
where $\theta_j$ is the angle between $\hat{x}$ and $\hat{W}_j$. To stabilize optimization and control logit scale, ArcFace introduces a scale factor $s>0$:
\[
  z_j = s \cdot \cos(\theta_j).
\]

\paragraph{Additive angular margin.} The key idea is to modify \emph{only} the correct class logit $y$ by adding an angular margin $m>0$ directly to the angle:
\[
  z_y = s \cdot \cos(\theta_y + m), \qquad z_{j \neq y} = s \cdot \cos(\theta_j).
\]
This choice yields a decision boundary with a clear geometric meaning on the hypersphere: to be classified correctly, a sample must be not only closer (in angular terms) to its class center, but closer by at least $m$. The final loss is standard cross-entropy applied to the modified logits:
\[
  \mathcal{L}_{\text{ArcFace}} = -\log \frac{e^{s\cos(\theta_y+m)}}{e^{s\cos(\theta_y+m)} + \sum_{j\neq y} e^{s\cos(\theta_j)}}.
\]
Compared to pair/triplet metric learning, this strategy exploits efficient supervised training over many classes/identities, while still producing an embedding usable outside the classification context.

\paragraph{Head} The head (weights $W_j$ and margin/scale) imposes geometric constraints during training. After training, in a biometric \emph{open identity} setting (unseen identities), the head is unusable because it is defined over training classes. Thus only the backbone is used to extract embeddings and vectors are compared via cosine similarity or distance (e.g., L2 on normalized vectors). In other words: training uses multi-class classification to learn a metric space; inference uses that space for verification/identification \cite{deng2019}.

\paragraph{Tabular MLP.} For tabular data we use an MLP backbone identical to the Siamese network with 13 fiducial inputs and 128-dimensional embeddings. The network is composed of fully-connected layers 32-64-256 with BatchNorm1d and ReLU, followed by dropout 0.10; the output is L2-normalized for retrieval and verification. 

\paragraph{Waveform CNN}
For waveforms we use a 1D CNN (Convolutional Neural Network) with 12-channel input. The network consists of 16 Conv1d blocks with BatchNorm and ReLU, kernels 7/24/16/8 and increasing dilations (up to 8), followed by dropout 0.05. Channels evolve in a 64-128-256-320 sequence, then 256 repeated and then 128 and 64; downsampling is obtained with MaxPool1d after blocks 4, 10, and 15. Output is aggregated by AdaptiveAvgPool1d and projected by a linear layer to a 512- or 1024-dimensional embedding, L2-normalized. A schematic is reported in Table~\ref{tab:waveform_v2}.

\section{Results}
\label{sec:risultati}

\subsection{Statistical results: INTRA vs INTER correlations}
\label{sec:risultati_stat}

On this temporal window (14,369 patients, 2 exams each) we observe a consistent separation between INTRA and INTER: Pearson INTRA $\sim$ 0.9964 vs INTER $\sim$ 0.9853 (gap $\sim$ +0.0111) and Spearman INTRA $\sim$ 0.9547 vs INTER $\sim$ 0.8951 (gap $\sim$ +0.0596), with rejection of $H_0$ in non-parametric tests and robustness verification via bootstrap/effect size. The evidence is consistent with the literature: separation can be moderate but systematic and remains detectable as temporal distance increases \cite{merone2017,pinto2018,melzi2023,meltzer2025,goldberger2000}.

On embedding representations (intra=7,239, inter=200,000) non-parametric tests confirm an even stronger separation: KS=0.965, AD statistic=27,626.91 (p<0.001), Mann-Whitney U=5,068,344 (p$\approx$0), Cramer-von Mises=2,290.96 (p=4.26e-07). Separation measures are extremely high (Cohen's d=5.29, Cliff's $\Delta$=0.993) and distribution overlap is low (Bhattacharyya=0.0647). The AUC computed on distances is 0.0035; reversing the score direction (inter vs intra) yields AUC$\approx$0.9965, consistent with the observed separation.

\subsection{Biometric results: embedding models}
\label{sec:risultati_embed}

The experimental flow shows a clear inflection point when moving from tabular inputs to waveform ECGs. The tabular stage uses only 13 fiducial variables (axes, durations, onset/offset), thus operating in a low-dimensional space; results are more modest but consistent with the signature hypothesis. Moving to waveforms increases information content and yields a sharp metric jump, in line with CNN-based literature on full signals \cite{lugovaya2005,odinaka2012,donida2019,altan2019,zhang2021,alduwaile2021}.

Normalization is a discriminative factor: train\_global (train\_norm) improves metrics relative to per\_sample (batch\_norm), suggesting better statistical alignment across train/val/test. The observed progression is: (i) baseline Siamese tabular, (ii) ArcFace tabular with a slight improvement, (iii) ArcFace on waveforms with the first major jump, (iv) a further jump when moving from batch\_norm to train\_norm at equal waveform setting, and (v) a final increase with dataset\_id=1001, which has more training examples than 1002.

Two normalization choices were tested:

\begin{itemize}
\item per\_sample: normalization per signal
\item train\_global: global statistics estimated on training

\end{itemize}
Normalization is often a determinant factor in biometric signals and deep learning on ECG, and the literature highlights that preprocessing differences contribute to result variability and transferability \cite{merone2017,pinto2018,melzi2023}.

Summary table of performance (verification + identification) across representations (Table~\ref{tab:perf_closed}).

\begin{table}[ht]
\centering
\caption{Closed-set performance on tabular features and waveforms (250/500/1000 Hz).}\label{tab:perf_closed}
\begin{tabular}{lrrrr}
\toprule
metric & tabular & waveform 250 & waveform 500 & waveform 1000 \\
\midrule
rank@1 & 0.350 & 0.651 & 0.776 & 0.812 \\
rank@5 & 0.622 & 0.800 & 0.868 & 0.888 \\
rank@10 & 0.720 & 0.844 & 0.893 & 0.910 \\
tar@far0.001 & 0.280 & 0.810 & 0.896 & 0.908 \\
tar@far0.0001 & 0.112 & 0.624 & 0.790 & 0.820 \\
\bottomrule
\end{tabular}
\end{table}

Performance refers respectively to tabular data, waveforms resampled at 250 Hz, 500 Hz, and original 1 kHz.

On the 1 kHz test set metrics are: TAR@FAR=0.001: 0.908 and TAR@FAR=0.0001: 0.820, with EER=0.0253 (2.53\%) at threshold 0.305. In identification we obtain Rank@1=0.812, Rank@5=0.888, Rank@10=0.910, and rank\_k\_95=59. In training, top-1 accuracy reaches $\sim$100\%, as expected for classification on seen identities.

In the open-set setting with gallery\_strategy = random\_single, the two-stage pipeline shows high DIR@FAR performance, as reported in Table~\ref{tab:dir_far_open_set}. In this evaluation, 5,000 gallery templates, 2,000 known probes, and 6,000 impostor probes were used. Thresholds are calibrated on impostors; DIR measures the fraction of known probes correctly identified and accepted at threshold.

\begin{table}[ht]
\centering
\small
\caption{Open-set: DIR@FAR (gallery\_strategy = random\_single).}\label{tab:dir_far_open_set}
\begin{tabular}{lcc}
\toprule
\textbf{strategy (stage 2)} & \textbf{DIR@FAR=1e-3} & \textbf{DIR@FAR=1e-4} \\
\midrule
bestofk & 0.976 & 0.976 \\
topkmean & 0.903 & 0.854 \\
s-norm & 0.976 & 0.975 \\
\bottomrule
\end{tabular}
\end{table}

Overall, bestofk and s-norm are highly effective in this setting, suggesting that the shortlist produced by stage 1 is sufficiently clean and that stage 2 mainly acts as a decision/calibration mechanism at fixed FAR.

\section{Discussion}

Much of the ECG biometrics literature operates on small to medium databases, often with heterogeneous protocols. Reviews and benchmarks (Pinto 2018; Melzi 2023; Meltzer 2025; Merone 2017) stress that differences in session, database, and acquisition conditions can dominate method comparisons and explain why apparently ``very high'' results on ECG-ID or PTB are not directly transferable to large-scale scenarios \cite{merone2017,pinto2018,melzi2023,goldberger2000}. Classic and comparative works (Israel 2005; Wang 2008; Odinaka 2012; Fratini 2015) demonstrate feasibility, but quantitative comparison depends on the number of subjects, leads, and protocols \cite{meltzer2025,lugovaya2005,wagner2020,donida2019}.

In the benchmark by Melzi et al., for example, PTB and ECG-ID have subject counts on the order of $10^2$, whereas their in-house dataset reaches tens of thousands and explicitly introduces a ``sinus rhythm only'' filter and single-session vs multi-session distinction \cite{pinto2018}. This is relevant because it shows that, already in literature, when scale increases and sessions vary, metrics worsen compared to controlled settings. The test set in this work lies in the large-scale range and makes the comparison more realistic.

\paragraph{EER in all-vs-all regime}

An EER of 2.53\% computed all-vs-all is particularly significant because:

\begin{itemize}
\item the number of impostor pairs grows very rapidly with population size;
\item the probability of ``hard'' impostors increases (subjects with similar physiological patterns);
\item the estimate does not depend on favorable impostor sub-sampling choices.

\end{itemize}
In many works, EER is computed on impostor pairs generated by sampling; in large-scale contexts this can yield optimistic values. The approach adopted here provides a robust and defensible estimate. Operationally, EER=2.53\% corresponds (at the balance point) to TAR $\sim$ 97.5\% and FAR $\sim$ 2.5\% with threshold $\sim$ 0.305. The fact that, at very low FAR, TAR remains high (0.908 at 1e-3; 0.820 at 1e-4) suggests that the tails of genuine/impostor distributions are already well separated, and that the TAR drop is the natural price of a more conservative threshold.

\paragraph{Verification}

In deployment contexts, security requirements often impose very low FAR. Thus FAR=1e-3 and FAR=1e-4 are the most informative metrics:

\begin{itemize}
\item TAR $\sim$ 0.908 @ FAR=1e-3
\item TAR $\sim$ 0.820 @ FAR=1e-4

\end{itemize}
These results are relevant because (i) they are measured on a large population, and (ii) they are operational: they directly indicate the fraction of genuine users retained when limiting false accept risk. This choice is consistent with standard biometric testing/reporting guidance and with the need, highlighted by reviews, to go beyond generic ``accuracy'' \cite{chee2022}.

\paragraph{Identification}

In the literature, identification accuracy (closed-set rank@1) is often reported as very high (sometimes >95-99\%) but on small databases, especially under controlled conditions or on beats/segments. For example, recent works report high results on PTB/ECG-ID using CNNs or more modern architectures, but on a few dozen or hundred subjects and different protocols \cite{odinaka2012,donida2019,altan2019,zhang2021,alduwaile2021}. The scenario considered here is more severe: 20k identities make identification a large-scale retrieval problem.

Beyond closed-set, open-set evaluation via DIR@FAR highlights the practical utility of a two-stage pipeline: stage 1 candidate generation on ArcFace embeddings and stage 2 re-ranking/decision on top-$K$. In this setting, high DIR@FAR values (up to 0.976 at FAR=1e-3 and 1e-4) indicate that shortlist + threshold decision can maintain high correct identification probability even under strict security constraints.

Rank@1=0.812 implies that in about 3 probes out of 4 the correct identity is at rank 1; Rank@10=0.910 indicates that in nearly 90\% of cases the correct identity is among the top 10 candidates. This has operational value: it enables two-stage pipelines (candidate generation $\rightarrow$ re-ranking/verification), a typical pattern for large watchlist systems. In particular, a high Rank@5 and a rank\_k\_95=59 (i.e., 95\% probability that the correct identity is among the top 59 candidates) make it natural to implement a second identification/verification level on top-$K$, reducing computational cost at fixed coverage.

Starting from an ArcFace embedding model, common strategies for stage two include: (i) \emph{re-ranking} with a more expressive measure (e.g., learned similarity or a Siamese/contrastive model) and/or a heavier waveform model; (ii) \emph{template aggregation} (mean/median embeddings per identity and/or per session) to reduce intra-subject noise; (iii) \emph{score-level fusion} across multiple scores (cosine on embeddings, per-lead score, quality score) with calibration; (iv) \emph{quality-aware re-weighting/gating} of comparisons (excluding or down-weighting low-quality signals).

\paragraph{Consistency between statistical motivation and deep performance: ``signature present and persistent''}

The initial statistical motivation (INTRA>INTER over windows up to multiple years) provides an interpretive bridge: the individual signature is observable already in tabular representations and remains visible as temporal distance increases (with mild attenuation). The literature confirms that ECG biometrics suffer from multi-session and temporal variability, but do not completely lose identity information; several works explicitly discuss long-term stability \cite{merone2017,pinto2018,melzi2023,goldberger2000}. In this sense, strong performance at very low FAR on a large-scale test set suggests that the learned embedding captures relatively stable components of cardiac identity, beyond potential demographic or physiological correlates.

Additional evidence comes from statistical analyses on embeddings: intra/inter distributions are strongly separated (KS=0.965; AD statistic=27,626.9; MW U=5,068,344; CvM=2,290.96, with extremely small p-values), with very large effect sizes (Cohen's d=5.29; Cliff's $\Delta$=0.993) and low overlap (Bhattacharyya=0.0647). Since the score is a distance, the AUC computed intra vs inter is inverted (0.0035), equivalent to AUC~0.9965 after inversion. These results reinforce the hypothesis of a persistent and coherent biometric signature in embedding space.

\subsection{Limitations and implications}

Two aspects deserve attention in the final writing:

\begin{enumerate}
\item Comparability with existing studies: it should be explicit that comparing ``raw numbers'' (accuracy/EER) across papers is difficult without aligning dataset, session, leads, preprocessing, and pairing protocol. This point is central in reviews \cite{merone2017,pinto2018,melzi2023}.

\item Definition of the target population: device-specific filtering and outlier cleaning make the cohort more homogeneous, improving interpretability, but delimit the domain of validity (e.g., other devices/centers). This should be presented as a methodological choice consistent with the goal: measuring an individual signature under controlled conditions and then extending.

\end{enumerate}

\subsection{Future work}

An immediate direction is testing on large external datasets such as the Harvard-Emory ECG Database (HEEDB), a clinical collection of 12-lead, 10 s ECGs sampled at 250/500 Hz, with tens of millions of traces and metadata/annotations (12SL, ICD) under controlled access \cite{koscova2025heedb,heedb_pmc}. This will enable assessment of transferability and robustness across populations and devices. Further developments include: (i) training and evaluation on single leads to identify the most informative lead; (ii) analysis of the most informative lead at fixed embedding (frozen network), to quantify how much identity information each lead retains.

\section{Conclusions}

In a large-scale setting ($\sim$23k identities; $\sim$56k ECGs), the system attains EER=2.53\% with all-vs-all estimation and maintains high performance at strict operating points (TAR=0.908 @ FAR=1e-3; TAR=0.820 @ FAR=1e-4). In identification, Rank@1=0.812 and Rank@10=0.910 indicate effective retrieval on a large population and support the use of two-stage pipelines. Overall, the results are consistent with the literature recognizing a persistent individual ECG signature and highlight the importance of operational metrics and robust protocols to move ECG biometrics toward realistic scenarios.

\clearpage
\hrule
\section*{Supplementary Tables}
\begin{table}[ht]
\centering
\small
\caption{Patient distribution by number of exams across the three refinement steps.}\label{tab:steps}
\begin{tabular}{lrrr}
\toprule
n. exams & Step 1 & Step 2 & Step 3 \\
\midrule
2 & 36,397 & 26,631 & 26,320 \\
3 & 18,578 & 12,468 & 12,200 \\
4 & 11,278 & 6,673 & 6,460 \\
5 & 7,259 & 3,716 & 3,601 \\
6 & 4,949 & 2,115 & 2,027 \\
7 & 3,398 & 1,287 & 1,234 \\
8 & 2,622 & 744 & 709 \\
9 & 1,974 & 430 & 397 \\
10 & 1,489 & 154 & 131 \\
11+ & 4,579 & 0 & 0 \\
\midrule
\textbf{Total} & \textbf{92,523} & \textbf{54,218} & \textbf{53,079} \\
\bottomrule
\end{tabular}
\end{table}

\begin{table}[ht]
\centering
\small
\caption{Waveform CNN v2: architectural summary (backbone).}\label{tab:waveform_v2}
\begin{tabular}{ll}
\toprule
Block & Main details \\
\midrule
Input & 12 channels (multi-lead ECG) \\
Conv1--2 & 64$\rightarrow$128 channels, kernel 7, dilation 1 \\
Conv3 & 256 channels, kernel 7, dilation 2 \\
Conv4 & 320 channels, kernel 24, dilation 1 + MaxPool \\
Conv5--8 & 256 channels, kernel 16, dilations 2/4/4/4 \\
Conv9--14 & 128 channels, kernel 8, dilations 4/6/6/6/8/8 \\
Conv15--16 & 64 channels, kernel 8, dilation 8 + MaxPool (Conv15) \\
Pooling & AdaptiveAvgPool1d (global) \\
Projection & Linear $\rightarrow$ embedding 512, L2 norm \\
\bottomrule
\end{tabular}
\end{table}

\begin{table}[ht]
\centering
\caption{Extended physiological ranges used for outlier cleaning (tabular features).}\label{tab:range}
\begin{tabular}{lrrl}
\toprule
feature & min & max & unit \\
\midrule
VentricularRate & 40 & 120 & bpm \\
PRInterval & 100 & 240 & ms \\
QRSDuration & 60 & 150 & ms \\
QTInterval & 300 & 500 & ms \\
QTCorrected & 300 & 500 & ms \\
PAxis & -30 & 90 & deg \\
RAxis & -90 & 120 & deg \\
TAxis & -30 & 120 & deg \\
POnset & 50 & 500 & ms \\
POffset & 150 & 550 & ms \\
QOnset & 200 & 650 & ms \\
QOffset & 300 & 750 & ms \\
TOffset & 500 & 1200 & ms \\
\bottomrule
\end{tabular}
\end{table}

\begin{table}[ht]
\centering
\small
\caption{Train set: size and exam-per-patient distribution.}\label{tab:train}
\begin{tabular}{lrrrrr}
\toprule
\shortstack{dataset} & \shortstack{number\\of exams} & \shortstack{number\\of patients} & \shortstack{min\\exams/patient} & \shortstack{mean\\exams/patient} & \shortstack{max\\exams/patient} \\
\midrule
1000 & 89,576 & 21,407 & 3 & 4.18 & 10 \\
1001 & 60,153 & 11,647 & 4 & 5.16 & 10 \\
1002 & 39,473 & 6,479 & 5 & 6.09 & 10 \\
\bottomrule
\end{tabular}
\end{table}

\begin{table}[ht]
\centering
\small
\caption{Validation set: size and exam-per-patient distribution.}\label{tab:val}
\begin{tabular}{lrrrrr}
\toprule
\shortstack{dataset} & \shortstack{number\\of exams} & \shortstack{number\\of patients} & \shortstack{min\\exams/patient} & \shortstack{mean\\exams/patient} & \shortstack{max\\exams/patient} \\
\midrule
1000 & 37,352 & 15,836 & 2 & 2.36 & 10 \\
1001 & 51,988 & 20,716 & 2 & 2.51 & 10 \\
1002 & 62,460 & 23,300 & 2 & 2.68 & 10 \\
\bottomrule
\end{tabular}
\end{table}

\begin{table}[ht]
\centering
\small
\caption{Test set: size and exam-per-patient distribution.}\label{tab:test}
\begin{tabular}{lrrrrr}
\toprule
\shortstack{dataset} & \shortstack{number\\of exams} & \shortstack{number\\of patients} & \shortstack{min\\exams/patient} & \shortstack{mean\\exams/patient} & \shortstack{max\\exams/patient} \\
\midrule
1000 & 37,512 & 15,836 & 2 & 2.37 & 10 \\
1001 & 52,299 & 20,716 & 2 & 2.52 & 10 \\
1002 & 62,507 & 23,300 & 2 & 2.68 & 10 \\
\bottomrule
\end{tabular}
\end{table}

\clearpage
\hrule
\bibliographystyle{plain}
\bibliography{references}
\end{document}